\documentclass[11pt,a4paper]{article}
\usepackage[hyperref]{emnlp2020}
\usepackage{times}
\usepackage{latexsym}
\usepackage{pgfplots}
\pgfplotsset{width=.9\columnwidth}
\usepackage{url}
\usepackage{makecell}
\usepackage{booktabs}
\usepackage{amsmath}
\usepackage[colorinlistoftodos]{todonotes}

\usepackage{microtype}

\aclfinalcopy 
\def\aclpaperid{2974}

\newcommand\ti[1]{\textit{#1}}
\newcommand\tf[1]{\textbf{#1}}
\newcommand\ttt[1]{\texttt{#1}}
\newcommand\mf[1]{\mathbf{#1}}

\date{}

\newcommand{\fair}{$^1$}
\newcommand{\epfl}{$^3$}
\newcommand{\ucl}{$^2$}
\newcommand{\fairucl}{$^{1,2}$}

\title{Scalable Zero-shot Entity Linking with Dense Entity Retrieval}

\author{Ledell Wu,\fair{} Fabio Petroni,\fair{} Martin Josifoski,\epfl{}\thanks{~~Work done during internship with Facebook.}~ Sebastian Riedel,\fairucl{}  Luke Zettlemoyer\fair{} \\
\fair{}Facebook AI Research\\
\texttt{\{ledell, fabiopetroni, sriedel, lsz\}@fb.com} \\
\ucl University College London \\
\epfl Ecole Polytechnique Federale de Lausanne \\
\texttt{ martin.josifoski@epfl.ch} \\
}

\begin{document}
\maketitle
\begin{abstract}

This paper introduces a conceptually simple, scalable, and highly effective BERT-based entity linking model, along with an extensive evaluation of its accuracy-speed trade-off. We present a two-stage zero-shot linking algorithm, where each entity is defined only by a short textual description. The first stage does retrieval in a dense space defined by a bi-encoder that independently embeds the mention context and the entity descriptions. Each candidate is then re-ranked with a cross-encoder, that concatenates the mention and entity text. Experiments demonstrate that this approach is state of the art on recent zero-shot benchmarks (6 point absolute gains) and also on more established non-zero-shot evaluations (e.g. TACKBP-2010), despite its relative simplicity (e.g. no explicit entity embeddings or manually engineered mention tables). We also show that bi-encoder linking is very fast with nearest neighbour search (e.g. linking with 5.9 million candidates in 2 milliseconds), and that much of the accuracy gain from the more expensive cross-encoder can be transferred to the bi-encoder via knowledge distillation. Our code and models are available at \url{https://github.com/facebookresearch/BLINK}.
\end{abstract}

\section{Introduction}

Scale is a key challenge for entity linking; there are millions of possible entities to consider for each mention. To efficiently filter or rank the candidates, existing methods use different sources of external information, including manually curated mention tables~\cite{ganea2017deep}, incoming Wikipedia link popularity~\cite{yamada2016joint}, and gold Wikipedia entity categories~\cite{gillick2019learning}. In this paper, we show that BERT-based models set new state-of-the-art performance levels for large scale entity linking when used in a zero shot setup, where there is no external knowledge and a short text description provides the only information we have for each entity. We also present an extensive evaluation of the accuracy-speed trade-off inherent to large pre-trained models, and show is possible to achieve very efficient linking with modest loss of accuracy.

\begin{figure*}[t]
    \centering
    \includegraphics[width=\linewidth]{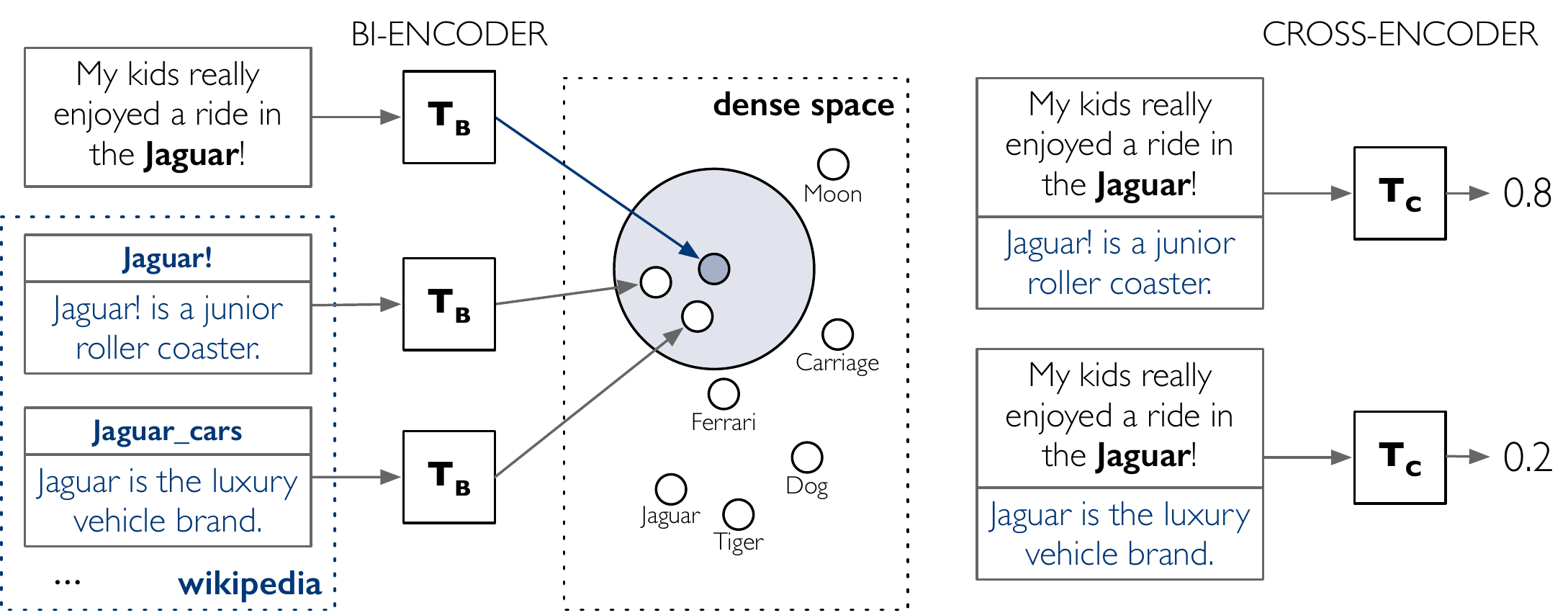}
    \caption{High level description of our zero-shot entity linking solution. From the top-left, the input gets encoded in the same dense space where all entities representations lie. A nearest neighbors search is then performed (depicted with a blue circle), $k$ entities retrieved and supplied to the cross encoder. The latter attends over both input text and entities descriptions to produce a probability distribution over the candidates. }
    \label{fig:high_level_desc}
\end{figure*}

More specifically, we introduce a two stage approach for zero-shot linking (see Figure \ref{fig:high_level_desc} for an overview), based on fine-tuned BERT architectures \cite{devlin2019bert}. In the first stage, we do retrieval in a dense space defined by a bi-encoder that independently embeds the mention context and the entity descriptions~\cite{humeau2019polyencoders, gillick2019learning}. 
Each retrieved candidate is then examined more carefully with a cross-encoder that concatenates the mention and entity text, following~\citet{logeswaran2019zero}. This overall approach is conceptually simple but highly effective, as we show through detailed experiments. 

Our two-stage approach achieves a new state-of-the-art result on TACKBP-2010, with an over 30\% relative error reduction. By simply reading the provided text descriptions, we are able to outperform previous methods that included many extra cues such as entity name dictionaries and link popularity.  We also improve the state of the art on existing zero-shot benchmarks, including a nearly 6 point absolute gain on the recently introduced Wikia corpus~\cite{logeswaran2019zero} and more than 7 point absolute gain on WikilinksNED Unseen-Mentions~\cite{onoe2019fine}. 

Finally, we do an extensive evaluation of the accuracy-speed trade-off inherent in our bi- and cross-encoder models. 
We show that the two stage methods scales well in a full Wikipedia setting, by linking against all the 5.9M Wikipedia entities for TACKBP-2010, while still outperforming existing model with much smaller candidate sets. 
We also show that bi-encoder linking is very fast with approximate nearest neighbor search (e.g. linking over 5.9 million candidates in 2 milliseconds), and that much of the accuracy gain from the more expensive cross-encoder can be transferred to the bi-encoder via knowledge distillation. 
We release our code and models, as well as a system to link entity mentions to all of Wikipedia (similar to TagME~\cite{ferragina2011fast}).\footnote{Our code and models are available at \url{https://github.com/facebookresearch/BLINK}}

\section{Related Work}
\label{sec:relatedwork}

We follow most recent work in studying entity linking with gold mentions.\footnote{\citet{kolitsas2018end} study end-to-end linking. Our techniques should be applicable to this setting as well, but we leave this exploration to future work.}
The entity linking task can be broken into two steps: candidate generation and ranking. Prior work has used frequency information, alias tables and TF-IDF-based methods for candidate generation. For candidate ranking, \newcite{he-etal-2013-learning}, \newcite{sun2015modeling}, \newcite{yamada2016joint}, \newcite{ganea2017deep}, and \newcite{kolitsas2018end} have established state-of-the-art results using neural networks to model context word, span and entity.  There is also recent work demonstrating that fine-grained entity typing information helps linking~\cite{raiman2018deeptype,onoe2019fine,khalife2018scalable}. 

Two recent results are most closely related to our work. \newcite{logeswaran2019zero} proposed the zero-shot entity linking task. They use cross-encoders for entity ranking, but rely on traditional IR-techniques for candidate generation and did not evaluate on large scale benchmarks such as TACKBP. 
\newcite{gillick2019learning} show that dense embeddings work well for candidate generation, but they did not do pre-training and included external category labels in their bi-encoder architectures, limiting their linking to entities in Wikipedia. Our approach can be seen as generalizing both of these lines of work, and showing for the first time that pre-trained zero-shot architectures are both highly accurate and computationally efficient at scale. 

\newcite{humeau2019polyencoders} studied different architectures to use deep pre-trained bidirectional transformers and performed detailed comparison of three different architectures, namely bi-encoder, poly-encoder, cross-encoder on tasks of sentence selection in dialogues. Inspired by their work, we use similar architectures to the problem of entity linking, and in addition, demonstrate that bi-encoder can be a strong model for retrieval. Instead of using the poly-encoder as a trade-off between cross-encoder and bi-encoder, we propose to train a bi-encoder model with knowledge distillation~\cite{model_compression, hinton2015distilling} from a cross-encoder model to further improve the bi-encoder's performances.

\section{Definition and Task Formulation}
\label{sec:definition}

\paragraph{Entity Linking}
Given an input text document $\mf{D} = \{ w_1, ..., w_r \}$ and a list of entity mentions $\mf{M_D} = \{ m_1, ..., m_n \}$, the output of an entity linking model is a list of mention-entity pairs $\{ (m_i, e_i) \}_{i \in [1, n]}$ 
where each entity is an entry in a knowledge base (KB) (e.g. Wikipedia), $e \in \mathcal{E}$. We assume that the title and description of the entities are available, which is a common setting in entity linking \cite{ganea2017deep,logeswaran2019zero}.
We also assume each mention has a valid gold entity in the KB, which is usually referred as \ti{in-KB} evaluation. We leave the out-of-KB prediction (i.e. \ti{nil} prediction) to future work.

\paragraph{Zero-shot Entity Linking}
We also study zero-shot entity linking~\cite{logeswaran2019zero}. Here the document setup is the same, but the knowledge base is separated in training and test time. Formally, denote $\mathcal{E}_{train}$ and $\mathcal{E}_{test}$ to be the knowledge base in training and test, we require $\mathcal{E}_{train}	\cap \mathcal{E}_{test} = \emptyset$. 
The set of text documents, mentions, and entity dictionary are separated in training and test so that the entities being linked at test time are unseen.
\section{Methodology} 
\label{sec:metod}

Figure~\ref{fig:high_level_desc} shows our overall approach. The bi-encoder uses two independent BERT transformers to encode model context/mention and entity into dense vectors, and each entity candidate is scored as the dot product of these vectors. The candidates retrieved by the bi-encoder are then passed to the cross-encoder for ranking. The cross-encoder encodes context/mention and entity in one transformer, and applies an additional linear layer to compute the final score for each pair. 

\subsection{Bi-encoder}
\label{sec:biencoder}
               
\paragraph{Architecture}
    We use a bi-encoder architecture similar to the work of \citet{humeau2019polyencoders} to model (mention, entity) pairs. This approach allows for fast, real-time inference, as the candidate representations can be cached. Both input context and candidate entity are encoded into vectors:
    \begin{align}
    \boldsymbol{y_{m}} = \mathrm{red}(T_1(\tau_m)) \\
    \boldsymbol{y_{e}} = \mathrm{red}(T_2(\tau_e))
    \end{align}
    where $\tau_m$ and $\tau_e$ are input representations of mention and entity respectively, $T_1$ and $T_2$ are two transformers. $\mathrm{red}(.)$ is a function that reduces the sequence of vectors produced by the transformers into one vector. Following the experiments in \citet{humeau2019polyencoders}, we choose $\mathrm{red}(.)$ to be the last layer of the output of the \ttt{[CLS]} token.
    
\paragraph{Context and Mention Modeling} 
    The representation of context and mention $\tau_m$ is composed of the word-pieces of the context surrounding the mention and the mention itself. Specifically, we construct input of each mention example as: 

\thinspace 
 
    \ttt{[CLS]} ctxt$_{l}$ \ttt{[M$_s$]} mention \ttt{[M$_{e}$]} ctxt$_{r}$ \ttt{[SEP]}
  
\thinspace

    \noindent where mention, ctxt$_{l}$, ctxt$_{r}$ are the word-pieces tokens of the mention, context before and after the mention respectively, and \ttt{[M$_{s}$]}, \ttt{[M$_{e}$]} are special tokens to tag the mention. The maximum length of the input representation is a hyperparameter in our model, and we find that small value such as 32 works well in practice (see Appendix A). 
    
 \paragraph{Entity Modeling}
    The entity representation $\tau_e$ is also composed of word-pieces of the entity title and description (for Wikipedia entities, we use the first ten sentences as description).
    The input to our entity model is:
    
\thinspace 

    \ttt{[CLS]} title \ttt{[ENT]} description \ttt{[SEP]}
    
\thinspace 

    \noindent where title, description are word-pieces tokens of entity title and description, and \ttt{[ENT]} is a special token to separate entity title and description representation.
    
\paragraph{Scoring} 
    The score of entity candidate $e_i$ is given by the dot-product:
    \begin{align}
    \label{eq:bi:sim}
    s(m, e_i) =  \boldsymbol{y_{m}} \cdot \boldsymbol{y_{e_i}} 
    \end{align}
    
\paragraph{Optimization}
    The network is trained to maximize the score of the correct entity with respect to the (randomly sampled) entities of the same batch~\cite{lerer2019pytorch,humeau2019polyencoders}. Concretely, for each training pair $(m_i, e_i)$ in a batch of $B$ pairs, the loss is computed as:
    \begin{align}
    \label{eq:bi:softmax}
    \mathcal{L}(m_i, e_i) = -s(m_i, e_i) + \log \sum_{j=1}^B \exp{(s(m_i, e_j))} 
    \end{align}
    
\citet{lerer2019pytorch} presented a detailed analysis on speed and memory efficiency of using batched random negatives in large-scale systems. In addition to in-batch negatives, we follow \citet{gillick2019learning} by using hard negatives in training. The hard negatives are obtained by finding the top 10 predicted entities for each training example. We add these extra hard negatives to the random in-batch negatives.

\paragraph{Inference}
At inference time, the entity representation for all the entity candidates can be pre-computed and cached. The inference task is then reduced to finding maximum dot product between mention representation and entity candidate representations. In Section~\ref{sec:exp:faiss} we present efficiency/accuracy trade-offs by exact and approximate nearest neighbor search using FAISS~\cite{johnson2019billion} in a large-scale setting. 

\subsection{Cross-encoder}
\label{sec:cross}
Our cross-encoder is similar to the ones described by \newcite{logeswaran2019zero} and \newcite{humeau2019polyencoders}. The input is the concatenation of the input context and mention representation and the entity representation described in Section~\ref{sec:biencoder} (we remove the \ttt{[CLS]} token from the entity representation). This allows the model to have deep cross attention between the context and entity descriptions.
Formally, we use $y_{m, e}$ to denote our context-candidate embedding:
\begin{align}
\boldsymbol{y_{m, e}} = \mathrm{red}(T_{\mathrm{cross}}(\tau_{m, e}))
\end{align}
where $\tau_{m,e}$ is the input representation of mention and entity, $T_{cross}$ is a transformer and $red(.)$ is the same function as defined in Section \ref{sec:biencoder}.

\paragraph{Scoring}
To score entity candidates, a linear layer $\boldsymbol{W}$ is applied to the embedding $\boldsymbol{y_{m, e}}$:
\begin{align}
s_{\mathrm{cross}}(m, e) = \boldsymbol{y_{m, e}} \boldsymbol{W}
\end{align}

\paragraph{Optimization}
Similar to methods in Section~\ref{sec:biencoder}, the network is trained using a softmax loss to maximize $s_{\mathrm{cross}}(m_i, e_i)$ for the correct entity, given a set of entity candidates (same as in Equation \ref{eq:bi:softmax}).

Due to its larger memory and compute footprint, we use the cross-encoder in a re-ranking stage, over a small set ($\le 100)$ of candidates retrieved with the bi-encoder. The cross-encoder is not suitable for retrieval or tasks that require fast inference.

\subsection{Knowledge Distillation}
\label{sec:kd}
To better optimize the accuracy-speed trade-off, we also report knowledge distillation experiments that use a cross-encoder as a teacher for a bi-encoder model. We follow \citet{hinton2015distilling} to use a softmax with temperature where the target distribution is based on the cross-encoder logits. 

Concretely, let $z$ be a vector of logits for set of entity candidates and $T$ a temperature, and $\sigma(z,T)$ a (tempered) distribution over the entities with  
\begin{align}
\sigma(z,T) = \frac{\exp{(z_i/T)}}{\sum_j \exp{(z_j/T)}}.
\end{align}
Then the overall loss function, incorporating both distillation and student losses, is calculated as 
\begin{align}
&\mathcal{L}_{dist} = \mathcal{H}(\sigma(z_t; \tau), \sigma(z_s; \tau)) \\
&\mathcal{L}_{st} =\mathcal{H}(e, \sigma(z_s; 1)) \\
&\mathcal{L} = \alpha \cdot \mathcal{L}_{st} + (1 - \alpha) \cdot \mathcal{L}_{dist}
\end{align}
where $e$ is the ground truth label distribution with probability 1 for the gold entity, $\mathcal{H}$ is the cross-entropy loss function, and $\alpha$ is coefficient for mixing distillation and student loss $\mathcal{L}_{st}$. The student logits $z_s$ are the output of the bi-encoder scoring function $s(m,e_i)$, the teacher logits the output of the cross-encoder scoring funcion $s_{\mathrm{cross}}(m, e)$. 
\section{Experiments}
\label{sec:evaluation}

In this section, we perform an empirical study of our model on three challenging datasets. 
\subsection{Datasets}
\label{sec:dataset}

\paragraph{The Zero-shot EL dataset} was constructed by \newcite{logeswaran2019zero} from Wikia.\footnote{\url{https://www.wikia.com.}} 
The task is to link entity mentions in text to an entity dictionary with provided entity descriptions, in a set of domains. There are 49K, 10K, and 10K examples in the train, validation, test sets respectively. 
The entities in the validation and test sets are from different domains than the train set, allowing for evaluation of performance on entirely unseen entities.  The entity dictionaries cover different domains and range in size from 10K to 100K entities.
 
\paragraph{TACKBP-2010} is widely used for evaluating entity linking systems~\newcite{ji2010overview}.\footnote{\url{https://tac.nist.gov}} Following prior work, we measure \ti{in-KB} accuracy (P@1). There are 1,074 and 1,020 annotated mention/entity pairs derived from 1,453 and 2,231 original news and web documents on training and evaluation dataset, respectively. All the entities are from the TAC Reference Knowledgebase which contains 818,741 entities with titles, descriptions and other meta info.

\paragraph{WikilinksNED Unseen-Mentions} was created by \citet{onoe2019fine} from the original WikilinksNED dataset~\cite{eshel2017named}, which contains a diverse set of ambiguous entities spanning a variety of domains. In the Unseen-Mentions version, no mentions in the validation and test sets appear in the training set. The train, validation and test sets contain 2.2M, 10K, and 10K examples respectively. In this setting, the definition of unseen-mentions is different from that in zero-shot entity linking: entities in the test set can be seen in the training set. However, in both definitions no (mention, entity) pairs from test set are observed in the training set. In the unseen-mentions test set, about $25\%$ of the entities appear in training set.

\subsection{Evaluation Setup and Results}
\label{sec:results}
We experiment with both BERT-base and BERT-large \cite{devlin2019bert} for our bi-encoders and cross-encoders. The details of training infrastructure and hyperparameters can be found in Appendix A.
All models are implemented in PyTorch\footnote{\url{https://pytorch.org}} and optimizied with Adam~\cite{Adam}.  We use (base) and (large) to indicate the version of our model where the underlying pretrained transformer model is BERT-base and BERT-large, respectively. 

\subsubsection{Zero-shot Entity Linking}
\label{sec:exp:zeshel}
First, we train our bi-encoder on the training set, initializing each encoder with pre-trained BERT base. Hyper-parameters are chosen based on Recall@64 on validation datase. For specifics, see Appendix A.2.
Our bi-encoder achieves much higher recall than BM25, as shown in Figure \ref{fig:retrieval}. 
Following \citet{logeswaran2019zero}, we use the top 64 retrieved candidates for the ranker, and we report Recall@64 on train, validation and test in Table~\ref{tab:zeshel-recall}.

\begin{table}
    \centering
    \resizebox{\linewidth}{!}{   
    \begin{tabular}{l c c c}
         \toprule
         \textbf{Method} &  \textbf{Train} & \textbf{Validation} & \textbf{Test} \\
         \midrule 
         BM25 &76.86 & 76.22 & 69.13 \\
         Ours (bi-encoder) & 93.12 & 91.44 & 82.06 \\
         \bottomrule
    \end{tabular}
    }
    \caption{Recall@64 (\%) on Zero-shot EL dataset, for the BM25 approach and our dense space bi-encoder based retrieval. Results on Train/Valideation/Test set reported.}
    \label{tab:zeshel-recall}
\end{table}
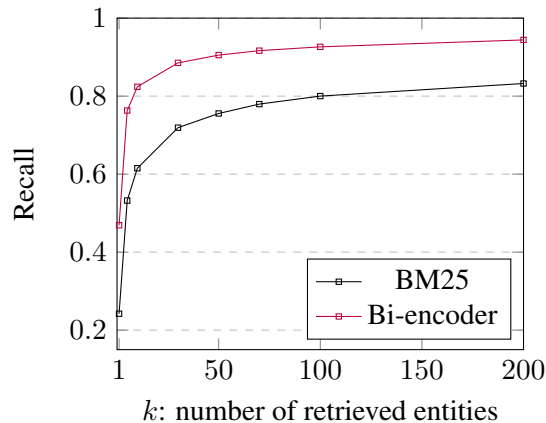
\begin{figure}[t!]
    \centering
        \begin{tikzpicture}
        \begin{axis}[
            xlabel={$k$: number of retrieved entities},
            ylabel={Recall},
            xmin=0, xmax=200,
            ymin=0.15, ymax=1,
            xtick={1,50,100,150,200},
            ytick={0,0.2,0.4,0.6,0.8,1},
            legend pos=south east,
            ymajorgrids=true,
            grid style=dashed,
        ]
         
        \addplot[
            color=black,
            mark=square,
            mark size=1pt
            ]
            coordinates {
            (1, 0.2422)(5, 0.5322)(10, 0.6151)(30, 0.7191)(50, 0.7555)(70, 0.7796)(100, 0.8001)(200, 0.8324)(300, 0.8498)(400, 0.8596)(500, 0.8651)
            };
            \addlegendentry{BM25}

        \addplot[
            color=purple,
            mark=square,
            mark size=1pt
            ]
            coordinates {
            (1,0.4689)(5,0.7631)(10,0.8240)(30,0.8853)(50,0.9052)(70,0.9166)(100,0.9264)(200,0.9441)(300,0.9530) (400,0.9580) (500,0.9617)
            };
            \addlegendentry{Bi-encoder}

        \end{axis}
        \end{tikzpicture}
    \caption{Top-$k$ entity retrieval recall on validation dataset of Zero-shot EL dataset}
    \label{fig:retrieval}
\end{figure}

After training the bi-encoder for candidate generation, we train our cross-encoder (initialized with pre-trained BERT) on the top 64 retrieved candidates from bi-encoder for each sample on the training set, and evaluate the cross-encoder on the test dataset. Overall, we are able to obtain a much better end-to-end accuracy, as shown in Table \ref{tab:zeshel}, largely due to the improvement on the retrieval stage.

\begin{table}[htb]
\centering
\begin{tabular}{l c}
    \toprule
    \textbf{Method} & \textbf{U.Acc.} \\
    \midrule
    \citet{logeswaran2019zero} & 55.08 \\
    \citet{logeswaran2019zero}(domain)$^\dagger$ & 56.58 \\
    \midrule
    Ours (base) & 61.34 \\
    Ours (large) & 63.03 \\
    \bottomrule
    \end{tabular}
\caption{\label{tab:zeshel} Performance on test domains on the Zero-shot EL dataset. U.Acc. represents the unnormalized accuracy. $\dagger$ indicates model trained with domain adaptive pre-training on source and target domain. Average performance across a set of worlds is computed by macro-averaging.}
\end{table}

We also report cross-encoder performance on the same retrieval method (BM25) used by \newcite{logeswaran2019zero} in Table~\ref{tab:zeshel-rank}, where the performance is evaluated on the subset of test instances for which the gold entity is among the top 64 candidates retrieved by BM25. We observe that our cross-encoder obtains slightly better results than reported by \newcite{logeswaran2019zero}, likely due to implementation and hyper-parameter details.

\begin{table}[htb]
\centering
    \begin{tabular}{l c c}
    \toprule
    \textbf{Method} & \textbf{Valid} & \textbf{Test} \\
    \midrule
    TF-IDF$^\dagger$ & 26.06 & \\
    \citet{ganea2017deep}$^\dagger$ & 26.96 & - \\
    \citet{gupta2017entity}$^\dagger$ & 27.03 & - \\
    \citet{logeswaran2019zero} & 76.06 & 75.06\\
    \midrule
    Ours (base) & 78.24 & 76.58\\
    \bottomrule
    \end{tabular}
\caption{\label{tab:zeshel-rank} Normalized accuracy on validation and test set on Zero-shot EL, where the performance is evaluated on the subset of test instances for which the gold entity is among the top-k candidates retrieved during candidate generation. $\dagger$ indicates methods re-implemented by \newcite{logeswaran2019zero}.}
\end{table}

\subsubsection{TACKBP-2010}
\label{sec:exp:tac}
Following prior work \cite{sun2015modeling, cao2018neural, gillick2019learning, onoe2019fine}, we pre-train our models on Wikipedia\footnote{\url{https://www.wikipedia.org/}} data. Data and model training details can be found in Appendix A.1. 

After training our model on Wikipedia, we fine-tune the model on the TACKBP-2010 training dataset. We use the top 100 candidates retrieved by the bi-encoder as training examples for the cross-encoder, and chose hyper-parameters based on cross validation. We report accuracy results in Table~\ref{tab:tackbp2010}. For ablation studies, we also report the following versions of our model:
\begin{enumerate}
    \item bi-encoder only: we use bi-encoder for candidate ranking instead of cross-encoder.
    \item Full Wikipedia:  we use 5.9M Wikipedia articles as our entity Knowlegebase, instead of TACKBP Reference Knowledgebase.
    \item Full Wikipedia w/o finetune: same as above, without fine-tuning on the TACKBP-2010 training set.
\end{enumerate}

As expected, the cross-encoder performs better than the bi-encoder on ranking. However, both models exceed state-of-the-art performance levels, demonstrating that the overall approach is highly effective. We observe that our model also performs well when we change the underlying Knowledgebase to full Wikipedia, and even without fine-tuning on the dataset. In Table \ref{tab:tac:retrieval} we show that our bi-encoder model is highly effective at retrieving relevant entities, where the underlying Knowledgebase is full Wikipedia.

\begin{table}
    \centering
    \begin{tabular}{l c}
    \toprule
         \textbf{Method} &  \textbf{Accuracy} \\
         \midrule
         \citet{he-etal-2013-learning} & 81.0 \\
         \citet{sun2015modeling} & 83.9 \\
         \citet{yamada2016joint}$^\dagger$  & 85.5 \\
         \citet{globerson2016collective}$^\dagger$ & 87.2 \\
         \citet{sil2018neural} & 87.4 \\
         \citet{nie2018mention}$^\dagger$  & 89.1 \\
         \citet{raiman2018deeptype} & 90.9 \\
         \citet{cao2018neural}$^\dagger$ & 91.0 \\
         \citet{gillick2019learning} & 87.0  \\
         \midrule
         Ours & 94.5 \\
         Ours (bi-encoder only) & 92.9 \\
         Ours (full Wiki) & 92.8 \\
         Ours (full Wiki, w/o finetune) & 91.5 \\
         \bottomrule

    \end{tabular}
    \caption{
    Accuracy scores of our proposed model and models from prior work on TACKBP-2010. $\dagger$ indicates methods doing $global$ resolution of all mentions in a document. Our work focuses on $local$ resolution where each mention is modeled independently.}
    \label{tab:tackbp2010}
\end{table}

There are however many other cues that could potentially be added in future work. For example, \newcite{khalife2018scalable} report $94.57\%$ precision on the TACKBP-2010 dataset. However, their method is based on the strong assumption that a gold fine-grained entity type is given for each mention (and they do not attempt to do entity type prediction). Indeed, if fine-grained entity type information is given by an oracle at test time, then \citet{raiman2018deeptype} reports $98.6\%$ accuracy on TACKBP-2010, indicating that improving fine-grained entity type prediction would likely improve entity linking. Our results is achieved without gold fine-grained entity type information. Instead, our model learns representations of context, mention and entities based on text only.

\begin{table}[t]
    \centering
        \begin{tabular}{l c}
        \toprule
        \textbf{Method} & \textbf{Recall@100}\\
        \midrule
        AT-Prior$^{\dagger} $& 89.5 \\
        AT-Ext$^{\dagger}$ & 91.7 \\
        BM25$^{\dagger}$ & 68.9 \\
        \citet{gillick2019learning} & 96.3 \\
           \midrule
        Ours (full wiki) & 98.3 \\
           \bottomrule
        \end{tabular}
    \caption{
    Retrieval evaluation comparison for TACKBP-2010. $\dagger$ indicates alias table and BM25 baselines implemented by \cite{gillick2019learning}. AT-Prior: alias table ordered by prior probabilities; AT-Ext: alias table extended with heuristics.}
    \label{tab:tac:retrieval}
\end{table}

\begin{table}[hbt]
    \centering
    \resizebox{\linewidth}{!}{  
    \begin{tabular}{l l c}
    \toprule
         \textbf{Method} &  \textbf{Training} & \textbf{Test} \\
         \midrule
         MOST FREQUENT & Wiki & 54.1 \\
         COSINE SIMILARITY & Wiki & 21.7 \\
         GRU+ATTN \\ \cite{mueller2018effective}  & in-domain & 41.2 \\
         GRU+ATTN & Wiki & 43.4 \\
         CBoW+WORD2VEC & in-domain & 43.0 \\
         CBoW+WORD2VEC & Wiki & 38.0 \\
         \citet{onoe2019fine} & Wiki & 62.2 \\
         \midrule
         Ours & in-domain & 74.7 \\
         Ours  & Wiki & 75.2 \\
         Ours  & Wiki (bi-encoder) & 71.5 \\
         Ours  & Wiki and in-domain & 76.8 \\
         \bottomrule
    \end{tabular}
    }
    \caption{Accuracy on the WikilinksNED Unseen-Mentions test set. The numbers of baseline models are from \cite{onoe2019fine}. The column \textbf{Training} indicates the source of data used in training: \textit{Wiki} means Wikipedia examples; \textit{in-domain} means examples in the training set.}
    \label{tab:wikilinks}
\end{table}

\subsubsection{WikilinksNED Unseen-Mentions}
Similarly to the approach described in Section \ref{sec:exp:tac}, we train our bi-encoder and cross-encoder model first on Wikipedia examples, then fine-tune on the training data from this dataset. We also present our model trained on Wikipedia examples and applied directly on the test set
as well as our model trained on this dataset directly without training on Wikipedia examples. We report our models' performance of accuracy on the test set in Table~\ref{tab:wikilinks}, along with baseline models presented from~\citet{onoe2019fine}. We observe that our model out-performs all the baseline models.

\paragraph{Inference time efficiency}
\label{sec:exp:faiss}
To illustrate the efficiency of our bi-encoder model, we profiled retrieval speed on a server with Intel Xeon CPU E5-2698 v4 @ 2.20GHz and 512GB memory. At inference time, we first compute all entity embeddings for the pool of 5.9M entities. This step is resource intensive but can be paralleled. On 8 Nvidia Volta v100 GPUs, it takes about 2.8 hours to compute all entity embeddings. Given a query of mention embedding, we use FAISS~\cite{johnson2019billion} IndexFlatIP index type (exact search) to obtain top 100 entity candidates. On the WikilinksNED Unseen-Mentions test dataset which contains 10K queries, it takes 9.2 ms on average to return top 100 candidates per query in batch mode.

We also explore the approximate search options using FAISS. We choose the IndexHNSWFlat index type following \citet{karpukhin2020dense}. It takes additional time in index construction while reduces the average time used per query. In Table~\ref{tab:faiss}, we see that $HNSW_1$\footnote{Neighbors to store per node: 128, construction time search depth: 200, search depth: 256; construction time: 2.1h.} reduces the average query time to 2.6 ms with less than 1.2\% drop in accuracy and recall, and $HNSW_2$\footnote{Neighbors to store per node: 128, construction time search depth: 200, search depth: 128; construction time: 1.8h.} further reduce the query time to 1.4 ms with less than 2.1\% drop.

\begin{table}[t]
    \centering
    \resizebox{\linewidth}{!}{  
    \begin{tabular}{l c c c c c c}
    \toprule
        \textbf{Method} & Acc & R@10 & R@30 & R@100 & ms/q \\
        \midrule
        Ex. Search & 71.5 & 92.7 & 95.4 & 96.7 & 9.2 \\
        $HNSW_1$ & 71.1 & 91.6 & 94.2 & 95.5 & 2.6 \\
        $HNSW_2$ & 70.7 & 91.0 & 93.9 & 94.6 & 1.4 \\
        \bottomrule
    \end{tabular}
   }
    \caption{Exact and approximate candidate retrieval using FAISS. Last column: average time per query (ms). }
    \label{tab:faiss}
\end{table}

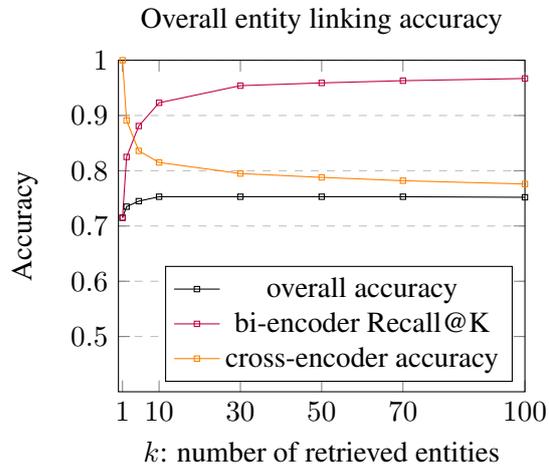
\begin{figure}[t!]
    \centering
        \begin{tikzpicture}
        \begin{axis}[
            title={Overall entity linking accuracy},
            xlabel={$k$: number of retrieved entities},
            ylabel={Accuracy},
            xmin=0, xmax=100,
            ymin=0.4, ymax=1,
            xtick={1,10,30,50,70,100},
            ytick={0,0.5,0.6,0.7,0.8,0.9,1},
            legend pos=south east,
            ymajorgrids=true,
            grid style=dashed,
        ]
         
        \addplot[
            color=black,
            mark=square,
            mark size=1pt
            ]
            coordinates {
            (1, 0.715)(2, 0.735)(5, 0.745)(10, 0.753)(30, 0.753)(50, 0.753)(70, 0.753)(100, 0.752)
            };
            \addlegendentry{overall accuracy}
            
        \addplot[
            color=purple,
            mark=square,
            mark size=1pt
            ]
            coordinates {
            (1, 0.715)(2, 0.825)(5, 0.881)(10, 0.923)(30,0.954)(50,0.959)(70,0.963)(100,0.967)
            };
            \addlegendentry{bi-encoder Recall@K}
            
        \addplot[
            color=orange,
            mark=square,
            mark size=1pt
            ]
            coordinates {
            (1, 1)(2, 0.891)(5, 0.836)(10, 0.815)(30, 0.795)(50,0.788)(70, 0.782)(100,0.776)
            };
            \addlegendentry{cross-encoder accuracy}

        \end{axis}
        \end{tikzpicture}
    \caption{Overall model accuracy based on different choices of $k$ (number of retrieved entities from biencoder), on the Unseen-Mentions dataset.}
    \label{fig:var_k}
\end{figure}

\paragraph{Influence of number of candidates retrieved}
In a two-stage entity linking systems, the choice of number of candidates retrieved influences the overall model performance. Prior work often used a fixed number of $k$ candidates where $k$ ranges from $5$ to $100$ (for instance, \citet{yamada2016joint} and \citet{ganea2017deep} choose $k=30$, \cite{logeswaran2019zero} choose $k=64$). When $k$ is larger, the recall accuracy increases, however, the ranking stage accuracy is likely to decrease. Further, increasing $k$ would often increase the run-time on the ranking stage. We explore different choices of $k$ in our model, and present the $recall@K$ curve, ranking stage accuracy and overall accuracy in Figure \ref{fig:var_k}. Based on the overall accuracy, we found that $k=10$ is optimal.

\begin{table*}[t!]
\centering
\resizebox{\textwidth}{!}{   

    \begin{tabular}{p{8cm} p{3.3cm} p{3.8cm}}
    \toprule
    \tf{Mention} & \tf{Bi-encoder} & \tf{Cross-encoder} \\
    \midrule
    But surely the biggest surprise is {\color{orange}\tf{Ronaldo}}’s drop in value, despite his impressive record of 53 goals and 14 assists in 75 appearances for Juventus. &
    \makecell[tl]{Ronaldo \\ (Brazilian footballer)} &
    \makecell[tl]{\tf{Cristiano Ronaldo}} \\
    
    \midrule
    ... they spent eleven days in the United Kingdom and Spain, photographing things like {\color{orange}\tf{Gothic}} statues, bricks, and stone pavements for use in textures. &
    \makecell[tl]{Gothic fiction} &
    \makecell[tl]{\tf{Gothic art}} \\
    
    \midrule
    To many people in many cultures, music is an important part of their way of life. {\color{orange}\tf{Ancient Greek}} and Indian philosophers defined music as tones ... &
    \makecell[tl]{\tf{Acient Greek}} &
    \makecell[tl]{Ancient Greek philosophy}\\
    
    
    \bottomrule

    \end{tabular}
}
    \caption{Examples of top entities predicted by Bi-encoder model and Cross-encoder model. Mentions in the examples are written in {\color{orange} \tf{ornage}} and the correct entity prediction in \tf{bold}.} 
\label{tab:retrieved-examples}
\end{table*}

\subsection{Knowledge Distillation}
In this section, we present results on knowledge distillation, using our cross-encoder as a teacher model and bi-encoder as a student model. 

We experiment knowledge distillation on the TACKBP-2010 and the WikilinksNED Unseen-Mentions dataset. 
We use the bi-encoder pretrained on Wikipedia as the student model, and fine-tune it on each dataset with knowledge distillation from the teacher model, which is the best performing cross-encoder model pretrained on Wikipedia and fine-tuned on the dataset. 

We also fine-tune the student model in our experiments on each dataset, without the knowledge distillation component, as baseline models.  
As we can see in Table \ref{tab:kd}, the bi-encoder model trained with knowledge distillation from cross-encoder out-performs the bi-encoder without knowledge distillation, providing another point in the accuracy-speed trade-off curve for these architectures. 

\begin{table}[hbt]
\centering
\resizebox{\linewidth}{!}{  
    \begin{tabular}{l c c c}
    \toprule
    \textbf{Dataset} & bi-encoder & teacher & bi-encoder-KD \\
    \midrule
    Unseen & 74.4 & 76.8 & 75.7 \\
    TAC2010 & 92.9 & 94.5 & 93.5 \\
    \bottomrule
    \end{tabular}
}
\caption{Knowledge Distillation Results. The teacher model is the cross-encoder, and bi-encoder-KD is the bi-encoder model trained with knowledge distillation. }
\label{tab:kd}
\end{table}
\section{Qualitative Analysis}
\label{set:qualitative}
Table~\ref{tab:retrieved-examples} presents some examples from our bi-encoder and cross-encoder model predictions, to provide intuition for how these two models consider context and mention for entity linking. 

In the first example, we see that the bi-encoder mistakenly links ``Ronaldo'' to the Brazilian football player, while the cross-encoder is able to use context word ``Juventus'' to disambiguate. In the second example, the cross-encoder is able to identify from context that the sentence is describing art instead of fiction, where the bi-encoder failed. In the third example, the bi-encoder is able to find the correct entity ``Ancient Greek,''; where the cross-encoder mistakenly links it to the entity ``Ancient Greek philosophy,'' likely because that the word ``philosophers'' is in context. We observe that cross-encoder is often better at utilizing context information than bi-encoder, but can sometimes make mistakes because of misleading context cues.

\section{Conclusion}
\label{sec:conclusion}

We proposed a conceptually simple, scalable, and highly effective two stage approach for entity linking. We show that our BERT-based model outperforms IR methods for entity retrieval, and achieved new state-of-the-art results on recently introduced zero-shot entity linking dataset, WikilinksNED Unseen-Mentions dataset, and the more established TACKBP-2010 benchmark, without any task-specific heuristics or external entity knowledge. We present evaluations of the accuracy-speed trade-off inherent to large pre-trained models, and show that it is possible to achieve efficient linking with modest loss of accuracy. 
Finally, we show that knowledge distillation can further improve bi-encoder model performance. 
Future work includes:
\begin{itemize}
\item Enriching entity representations by adding entity type and entity graph information;
\item Modeling coherence by jointly resolving mentions in a document;
\item Extending our work to other languages and other domains;
\item Joint models for mention detection and entity linking.
\end{itemize}

\section*{Acknowledgements}
We thank our colleagues Marjan Ghazvininejad, Kurt Shuster, Terra Blevins, Wen-tau Yih and Jason Weston for fruitful discussions.

\bibliographystyle{acl_natbib}
\bibliography{reference}

\newpage
\appendix
\section{Training details and hyper-parameters Optimization}
\label{appendix:overall}
\begin{itemize}
    \item Computing infrastructure: we use 8 Nvidia Volta v100 GPUs for model training.
    \item Bounds for each hyper parameter: see Table \ref{tab:bounds}. In addition, for our bi-encoders, we use a max number of tokens of $[32, 64, 128]$ for context/mention encoder and $128$ for candidate encoder. In our knowledge distillation experiments, we set $\alpha=0.5$, and $T$ in $[2, 5]$. We use grid search for hyperparameters, for a total number of $24$ trials. 
    \item Number of model parameters: see Table \ref{tab:num_params}.
    \item For all our experiments we use accuracy on validation set as criterion for selecting hyperparameters.
\end{itemize}

\begin{table}[h]
    \centering
    \resizebox{\linewidth}{!}{   
    \begin{tabular}{l l}
         \toprule
         \textbf{Parameter} &  Bounds \\
         \midrule 
         Learning rate & [$2\mathrm{e}^{-6}$, $5\mathrm{e}^{-6}$, $1\mathrm{e}^{-5}$, $2\mathrm{e}^{-5}$] \\
         Bi-encoder batch size & $[128, 256]$ \\
         Cross-encoder batch size & $[1, 5]$ \\
         \bottomrule
    \end{tabular}
    }
    \caption{Bounds of hyper-parameters in our models}
    \label{tab:bounds}
\end{table}

\begin{table}[h]
    \centering
    \resizebox{\linewidth}{!}{   
    \begin{tabular}{l c}
         \toprule
         \textbf{Model} &  Number of parameters \\
         \midrule 
         Bi-encoder (base) & 220M \\
         Cross-encoder (base) & 110M \\
         Bi-encoder (large) & 680M \\
         Cross-encoder (large) & 340M \\
         \bottomrule
    \end{tabular}
    }
    \caption{Number of parameters in our models}
    \label{tab:num_params}
\end{table}


\subsection{Training on Wikipedia data}
\label{appendix:wiki}
We use Wikipedia data to train our models first, then fine-tune it on specific dataset. This approach is used in our experiments on TACKBP-2010 and WikilinksNED Unseen-Mentions datasets.

We use the May 2019 English Wikipedia dump which includes 5.9M entities, and use the hyperlinks in articles as examples (the anchor text is the mention). We use a subset of all Wikipedia linked mentions as our training data for the bi-encoder model (A total of 9M examples). We use a holdout set of 10K examples for validation. 
We train our cross-encoder model based on the top 100 retrieved results from our bi-encoder model on Wikipedia data. For the training of the cross-encoder model, we further down-sample our training data to obtain a training set of 1M examples. 

\paragraph{Bi-encoder (large) model}
Hyperparameter configuration for best model: learning rate=$1\mathrm{e}^{-5}$, batch size=128, max context tokens=32.
Average runtime for each epoch: 17.5 hours/epoch, trained on 4 epochs.

\paragraph{Cross-encoder (large) model}
Hyperparameter configuration for best model: learning rate=$2\mathrm{e}^{-5}$, batch size=1, max context tokens=32. Average runtime for each epoch: 37.2 hours/epoch, trained on 1 epoch.


\subsection{Zero-shot Entity Linking Dataset}
Dataset available at \url{https://github.com/lajanugen/zeshel}.
\label{appendix:zeshel}
There are 49K, 10K, and 10K examples in the train, validation, test sets respectively.
Training details: 
\paragraph{Bi-encoder (base) model} 
Hyperparameter configuration for best model: learning rate=$2\mathrm{e}^{-5}$, batch size=128, max context tokens=128.
Average runtime: 28.2 minutes/epoch, trained on 5 epochs.

\paragraph{Bi-encoder (large) model}
Hyperparameter configuration for best model: learning rate=$1\mathrm{e}^{-5}$, batch size=128, max context tokens=128.
Average runtime: 38.2 minutes/epoch, trained on 5 epochs.

\paragraph{Cross-encoder (base) model}
Hyperparameter configuration for best model: learning rate=$1\mathrm{e}^{-5}$, batch size=1, max context tokens=128.
Average runtime: 2.6 hours/epoch, trained on 2 epochs.

\paragraph{Cross-encoder (large) model}
Hyperparameter configuration for best model: learning rate=$1\mathrm{e}^{-5}$, batch size=1, max context tokens=128.
Average runtime: 8.5 hours/epoch, trained on 2 epochs.


\subsection{TACKBP-2010 Dataset}
\label{appendix:tac}
Dataset available at \url{https://catalog.ldc.upenn.edu/LDC2018T16}. There are 1,074 and 1,020 annotated examples in the train and test sets respectively. We use a 10-fold cross-validation from training set. Training details:

\paragraph{Bi-encoder (large) model}
Hyperparameter configuration for best model: learning rate=$2\mathrm{e}^{-6}$, batch size=128, max context tokens=32.
Average runtime: 9.0 minutes/epoch, trained on 10 epochs.
\paragraph{Bi-encoder (large) model with Knowledge Distillation}
Hyperparameter configuration for best model: learning rate=$2\mathrm{e}^{-5}$, batch size=128, max context tokens=32, $T=2, \alpha=0.5$.
Average runtime: 11.2 minutes/epoch, trained on 10 epochs.
\paragraph{Cross-encoder (large) model}
Hyperparameter configuration for best model: learning rate=$1\mathrm{e}^{-5}$, batch size=1, max context tokens=128.
Average runtime: 20.4 minutes/epoch, trained on 10 epochs.


\subsection{WikilinksNED Unseen-Mentions Dataset}
\label{appendix:unseen}
The train, validation and test sets contain 2.2M, 10K, and 10K examples respectively. We use a subset of 100K examples to fine-tune our model on this dataset, as we found more examples do not help. Training details:
\paragraph{Bi-encoder (large) model}
Hyperparameter configuration for best model: learning rate=$2\mathrm{e}^{-6}$, batch size=128, max context tokens=32.
Average runtime for each epoch: 3.2 hours/epoch, trained on 1 epochs.
\paragraph{Bi-encoder (large) model with Knowledge Distillation}
Hyperparameter configuration for best model: learning rate=$5\mathrm{e}^{-6}$, batch size=128, max context tokens=32, $T=2, \alpha=0.5$.
Average runtime: 6.5 hours/epoch, trained on 1 epochs.
\paragraph{Cross-encoder (large) model}
Hyperparameter configuration for best model: learning rate=$2\mathrm{e}^{-6}$, batch size=5, max context tokens=128.
Average runtime: 4.2 hours/epoch, trained on 1 epochs.


\end{document}